\documentclass[11pt]{article}

\usepackage[final]{acl}
\usepackage[skins,breakable]{tcolorbox}

\newtcolorbox{datasetbox}[1]{
  enhanced,
  colback=orange!5,            
  colframe=orange!20,          
  coltitle=black,            
  fonttitle=\rmfamily\bfseries\large,
  title=#1,                  
  arc=2.5mm,                 
  boxrule=1pt,               
  left=2.5mm, right=2.5mm, top=1.5mm, bottom=1.5mm,
  fontupper=\ttfamily\normalsize  
}
\usepackage{times}
\usepackage{latexsym}

\usepackage[T1]{fontenc}

\usepackage[utf8]{inputenc}

\usepackage{microtype}

\usepackage{inconsolata}
\usepackage{enumitem}

\usepackage{graphicx}
\usepackage{amsmath}
\usepackage{amssymb}
\usepackage{booktabs}
\usepackage{multirow}
\usepackage{cleveref}
\usepackage{bm}
\usepackage{colortbl}
\usepackage{xurl}
\usepackage{array}
\usepackage{xcolor}

\newcommand{\method}{VITAL}

\title{Learning When to Think Aloud in Latent Reasoning}

\author{
  \textbf{Zhehao Zhou\textsuperscript{1}},
  \textbf{Tongrui Su\textsuperscript{2}},
  \textbf{Shengyu Zhu\textsuperscript{2}}\thanks{Corresponding author.},
  \textbf{Da Li\textsuperscript{2,3}},
  \textbf{Zeyu Liu\textsuperscript{3}},
  \textbf{Jun Zhu\textsuperscript{3}},
  \textbf{Xueqi Cheng\textsuperscript{2}}
\\
\\
  \textsuperscript{1}Beijing Institute of Technology
\\
  \textsuperscript{2}Key Laboratory of AI Safety, Institute of Computing Technology, Chinese Academy of Sciences
\\
  \textsuperscript{3}Kuaishou Technology
\\
\\
  \texttt{1120232721@bit.edu.cn},
  \texttt{sutongrui25s@ict.ac.cn},
  \texttt{zhushengyu@ict.ac.cn}
}

\begin{document}
\maketitle
\begin{abstract}
Latent Chain-of-Thought (CoT) reasoning improves  efficiency
by shifting intermediate computation into  continuous hidden
states. However, fully latent methods often struggle on precision-critical tasks such as mathematical reasoning, while existing hybrid approaches retain substantial token-generation overhead. 
We introduce \method{}, an importance-guided hybrid reasoning framework that
selectively verbalizes informative tokens as reasoning anchors while keeping most computation latent. A lightweight gating head dynamically selects between
explicit token generation and direct hidden-state feedback. \method{} constructs hybrid  trajectories from explicit CoT using token-importance estimation and trains the model through curriculum-based supervised training, followed by reinforcement learning with a reward that balances correctness and conciseness. 
Experiments show that \method{} consistently outperforms existing fully latent reasoning methods, achieving relative average accuracy gains of 9.6\% on GPT-2 and 5.1\% on Llama-3.2-1B  at comparable reasoning lengths. Compared with hybrid baselines, \method{} achieves higher reasoning accuracy with substantially shorter reasoning trajectories on Qwen3-1.7B.
\end{abstract}

\section{Introduction}
\label{sec:intro}



Large Language Models (LLMs) have demonstrated remarkable
reasoning capabilities through Chain-of-Thought (CoT)
prompting, which decomposes complex problems into
step-by-step natural language rationales
\citep{wei2022chain}.
However, generating lengthy reasoning
traces incurs  substantial  decoding costs.
To improve efficiency, latent CoT methods perform
intermediate computation in continuous representations
rather than verbalizing every reasoning step.
Coconut \citep{hao2024coconut} introduces hidden-state
feedback for continuous reasoning, while CODI
\citep{shen2025codi} and CoLaR \citep{tan2025colar}
explore compressing explicit CoT into latent representations.
SIM-CoT \citep{wei2025sim} further applies step-level
supervision to latent reasoning, while ACoT \citep{ramji2026thinking}
reasons with discrete abstract tokens instead of continuous states.
Recent hybrid approaches, such as Token Assorted
\citep{su2025mix}, HRPO \citep{yue2026hybrid} and HybridCoT \citep{shen2026hybridcot}
combine latent and explicit representations to improve the overall reasoning performance.

Despite these advances, fully latent methods often underperform explicit CoT on precision-critical tasks like mathematical reasoning, where reliable multi-step computation requires a more accurate
handling of intermediate quantities and operations. \citet{wei2025sim} show that increasing latent depth can destabilize training and reduce step-level semantic diversity. \citet{zou2026capacityandlimits} introduce a notion of symbolic index to  explain the fundamental trade-off between exploration and execution in the reasoning process. Meanwhile, hybrid reasoning methods incorporate explicit intermediate tokens to improve the reasoning accuracy, but generally incur a greater token overhead than fully latent methods. For example, Token Assorted compresses only the initial portion of the reasoning trajectory, whereas HRPO
still generates a token at each reasoning step despite incorporating latent representations.
Based on these observations, we ask: \emph{Can we selectively verbalize a small number of informative tokens as discrete reasoning anchors to preserve critical intermediate information while keeping most computation latent?}  For training, these tokens serve as  explicit  prediction targets; during inference, their embeddings are fed back into the model to support subsequent latent computation.

\begin{figure*}[t]
    \centering
    \includegraphics[width=0.85\textwidth]{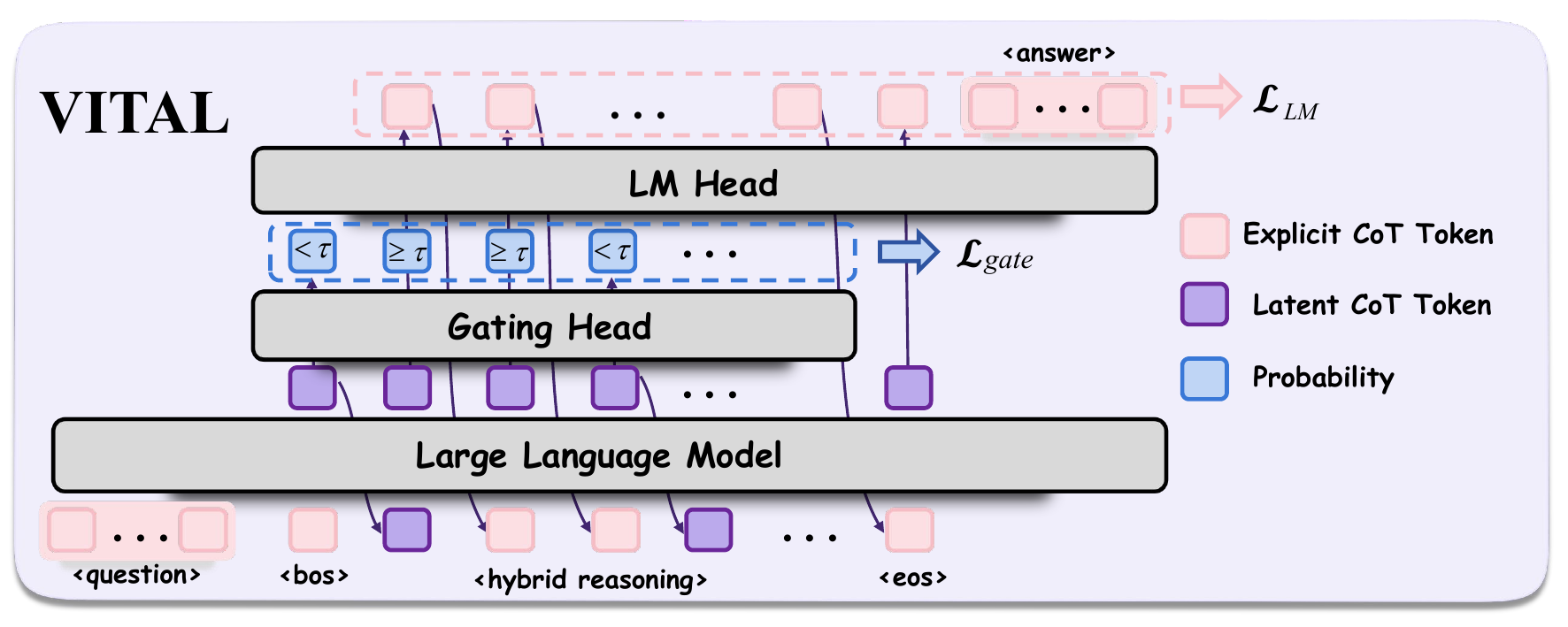}
    \caption{
    {Architecture of \method.}
    A lightweight gating head dynamically selects between explicit token generation
    and latent computation at each reasoning step.
    }
    \label{fig:framework}
\end{figure*}

In this work, we propose \method{}, a hybrid
reasoning framework that learns when to think aloud by
Verbalizing Informative Tokens as
 {reasoning Anchors} for Latent reasoning.
Rather than generating the full CoT, \method{} selectively
externalizes key intermediate tokens while keeping less
essential computation latent.
As illustrated in \Cref{fig:framework},
\method{} augments a base language model with a lightweight
gating head that dynamically decides whether to generate
an explicit token or continue latent computation.
In the former case, the model generates a token through
standard autoregressive decoding and feeds its embedding
back as the next input.
In the latter case, the current hidden state is directly
fed back for subsequent computation.
This mechanism enables \method{} to interleave explicit
reasoning anchors with continuous latent computation
within a unified autoregressive trajectory.

To train \method{}, we construct hybrid
trajectories from explicit CoT demonstrations.
Specifically, we employ LLMLingua-2
\citep{pan-etal-2024-llmlingua} to estimate token
importance within each reasoning step, retaining
high-importance tokens as explicit anchors while
compressing the remaining spans into latent representation.
We  adopt curriculum-based supervised training
that progressively transforms explicit CoT trajectories
into hybrid trajectories, and  further apply Reinforcement Learning (RL)
post-training with a tailored reward that balances both correctness and conciseness.

We evaluate \method{} on mathematical and scientific reasoning benchmarks using GPT-2 \citep{radford2019gpt2}, Llama-3.2-1B \citep{meta2024llama32}, and Qwen3-1.7B \cite{qwen3technicalreport}. Experimental results show that \method{} consistently outperforms existing fully latent reasoning baselines, achieving relative improvements in average accuracy of 9.6\% on GPT-2 and 5.1\% on Llama-3.2-1B at comparable reasoning lengths. Compared with hybrid methods  Token Assorted and HRPO, \method{} achieves higher reasoning accuracy with substantially shorter reasoning trajectories. Ablation studies further show that selecting  high-importance tokens achieves higher accuracy   than random or low-importance tokens. 
These findings validate the effectiveness of importance-guided selective verbalization in improving the accuracy-length trade-off of latent reasoning.

Our contributions are threefold:
\begin{itemize}[leftmargin=*,itemsep=-3pt,topsep=0pt]
    \item We propose \method{}, a hybrid reasoning framework that uses a
    lightweight gating head to interleave sparse explicit reasoning anchors
    with continuous latent computation.
    \item We develop an importance-guided, step-wise trajectory construction strategy
    together with a two-stage training procedure to enable selective verbalization of informative tokens.
    \item We evaluate \method{} against explicit, latent, and hybrid
    reasoning baselines across three backbones, demonstrating an improved accuracy-length trade-off and the importance of selecting informative reasoning anchors.
\end{itemize}

\section{Related Work}

\paragraph{Latent CoT Reasoning.}
Latent CoT methods seek to reduce explicit token generation by internalizing intermediate reasoning into continuous hidden states. Early work, such as iCoT \citep{deng2024explicit}, progressively internalizes explicit reasoning steps during training. Coconut \citep{hao2024coconut} learns continuous reasoning through curriculum-based hidden-state feedback, while CODI \citep{shen2025codi} compresses explicit CoT through self-distillation. CoLaR \citep{tan2025colar} learns to predict compressed embeddings with adjustable compression factors. SIM-CoT \citep{wei2025sim} introduces step-level supervision through an auxiliary decoder that is removed at inference. Unlike these approaches,  \method{} selectively verbalizes informative intermediate tokens as discrete reasoning anchors during inference. These tokens are then fed back into the model as embeddings to support subsequent latent computation.

\paragraph{Hybrid CoT Reasoning.}
Hybrid methods combine latent and explicit reasoning tokens. Token
Assorted \citep{su2025mix} replaces a prefix of the reasoning trace
with discrete latent codes, and HRPO \citep{yue2026hybrid} uses RL to
learn a gated blend of token embeddings and continuous representations
while still sampling a token at every step. HybridCoT
\citep{shen2026hybridcot} generates each text block in full, compresses
it into latent tokens, and keeps only LLM-annotated math spans in the
context, reducing context length rather than decoding steps.
In contrast, \method{} retains importance-selected tokens within every
reasoning step as anchors and never decodes the remaining content as
text, replacing it with latent computation. During inference, a
lightweight gating head selects between explicit token generation and
direct hidden-state feedback at each position. This design 
improves  reasoning accuracy while keeping the decoding trajectory short.

\paragraph{Soft-Embedding Reasoning.}
Another line of work performs continuous reasoning by constructing soft
embeddings from vocabulary distributions. Soft Thinking
\citep{zhang2026soft} replaces discrete token selection with a
probability-weighted mixture of vocabulary embeddings. SwiReasoning
\citep{shi2026swireasoning} and SeLaR \citep{fu2026selar} dynamically
switch between such soft embeddings and explicit token generation using
confidence signals. These approaches operate at inference without
additional training. In contrast, \method{} directly feeds back the
current hidden state during latent computation, rather than
constructing soft embeddings from vocabulary distributions, and learns
its routing policy from hybrid trajectories constructed from explicit
CoT demonstrations.


\section{Preliminary}
\label{sec:preliminary}
This section introduces the notation and the  token importance estimation method used in the paper.
\subsection{Notation}
We denote the base causal language model by $\mathcal{M}_\theta$. For an input sequence ${y}_{\le t}=(y_1,y_2,\cdots,y_t)$,  let
$
E_t = [e(y_1), e(y_2), \cdots, e(y_t)]$
be the sequence of input embeddings  up to position $t$, where $e(\cdot)$ represents the token embedding function. We also use $e_t$ to represent the input embedding at position $t$. The model produces hidden states
\begin{align}
\label{eqn:modelstate}
H_t = \mathcal{M}_\theta(E_t), \quad h_t = H_t[t,:],
\end{align}
where $H_t \in \mathbb{R}^{t \times d}$, and $h_t \in \mathbb{R}^d$ is the hidden state at the last position. Standard next-token prediction is given by
$
p_\theta(y_{t+1}\mid y_{\le t}) = \mathrm{Softmax}(W h_t),
$
with $W$ the language modeling head.

For explicit CoT output, we assume that each  example consists of a question $x$, a step-wise rationale $c=(s_1,\dots,s_N)$,   and a final response~$a$. Here  $s_i$ is the $i$-th reasoning step with tokens $(s_{i,1},s_{i,2},\cdots,s_{i,N_i})$.

\subsection{Token Importance}
\label{sec:tokenimportance}
Not every token in the CoT output contributes the same to the response. Recently,  \citet{li-etal-2023-compressing}  measure the importance of output tokens  using the semantic  confidence of LLMs: $-\log p_{\phi}(y_t\mid y_{<t})$ where  $\mathcal{M}_{\phi}$ is the LLM used to compute the confidence and  $\{y_i\}_{i=1}^t$ is the given tokens.

Despite its simplicity, \citet{pan-etal-2024-llmlingua} argue that there are two limitations. First, perplexity-based scores often assign lower importance to tokens near the end of a sentence. Second, causal language models rely on unidirectional attention and may fail to capture important contextual information.  Authors then introduce  LLMLingua-2 using a bidirectional BERT-like model \cite{devlin-etal-2019-bert} to estimate token importance. It utilizes GPT-4  to label
 whether a token is important, and  trains the bidirectional model with a  classification objective.
The importance is  measured by the predicted probability of each token: $p_{\phi'}(y_i\mid y_{\leq t})$ with $\mathcal{M}_{\phi'}$ the bidirectional model. We will treat LLMLingua-2 as a frozen scoring module and use the scores to decide which  tokens serve as  reasoning anchors. 

\section{\method{}: Learning When to Think Aloud}

This section presents \method{}, a hybrid autoregressive reasoning framework that combines explicit symbolic reasoning and latent computation. 
\begin{figure*}[t]
    \centering
    \includegraphics[width=0.9\textwidth]{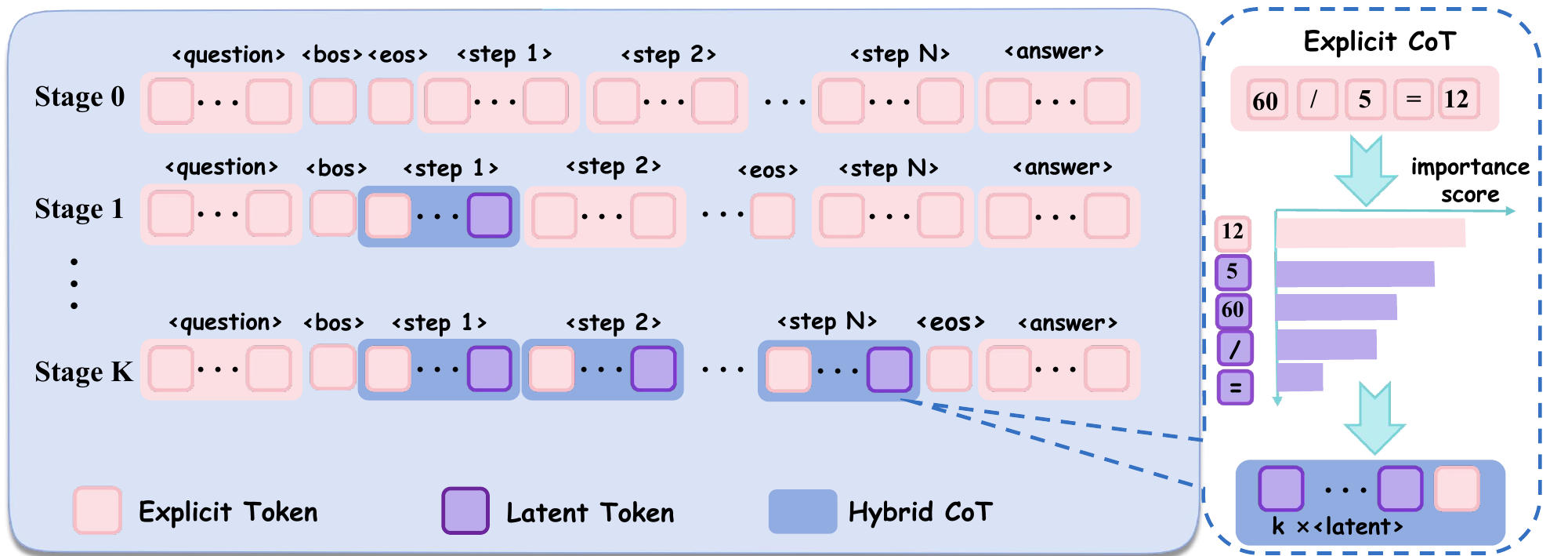}
    \caption{Curriculum-based supervised training for \method{}. We construct hybrid reasoning trajectories from explicit CoT by identifying high-importance
tokens as the  anchors.  }
    \label{fig:VOICEtrain}
\end{figure*}
 \subsection{Overview}
 \label{sec:method_overview}

 \method{} modifies a base language model  with a lightweight gating module, enabling it to operate in a mixed symbolic-latent space. The model dynamically switches between  generating explicit token  in the language space or performing latent computation in the continuous hidden space.

 \Cref{fig:VOICEtrain} illustrates the  training process of \method{}. We construct  hybrid trajectories from explicit rationales by identifying high-importance tokens as reasoning anchors.   We  train \method{} with    a joint objective consisting of language modeling loss and  token-wise gating loss, following a stage-wise curriculum-based procedure. Finally, we apply RL post-training to further improve the performance. 


\subsection{Hybrid Trajectory Construction}
\label{sec:hybridtraj}




\method{} employs LLMLingua-2~\citep{pan-etal-2024-llmlingua} introduced in \Cref{sec:tokenimportance} as an off-the-shelf token importance scoring model. Different from \citet{xia2025tokenskip}, we perform  token selection  individually within each reasoning step so that every step retains some symbolic anchors. This strategy helps preserve local symbolic structure. In contrast, selection over the entire CoT often results in a subset of tokens from merely one or a few reasoning steps, leading to degraded performance.

\paragraph{High-Importance Tokens.} Concretely, for each  step $s_i=\{s_{i,j}\}_j$, we use $\{q_{i,j}\}_j$ to denote corresponding importance scores. For our purpose,  we pick the top-$\gamma$, $\gamma\in(0,1)$ fraction of tokens in the reasoning step according to the importance score. That is, we rank the scores $\{q_{i,j}\}_{j}$ in a descending order, and denote by $q_i(\gamma)$  the empirical $\gamma$-quantile of these
importance values. Tokens with importance values  greater than or equal to $q_i(\gamma)$ are selected as anchor tokens: 
 \begin{align}
     \label{eqn:importtokens}
 \mathcal{S}_i = \left\{s_{i,j} \mid q_{i,j} \geq q_i(\gamma)\right\}.
\end{align}
The rest tokens are considered to be processed through latent computation.

\paragraph{Hybrid Trajectory.}Next, we convert each reasoning step into a mixed symbolic-latent form.  In particular,  consecutive ``unimportant'' tokens, i.e., tokens not in $\mathcal{S}_i$,  are compressed into a fixed number $k_{\text{latent}}$ of special \texttt{<latent>} placeholders, regardless of the original length. The  tokens in $\mathcal{S}_i$ are retained as ordinary tokens.   This construction preserves important tokens as anchors, while less essential  content is compressed into compact placeholder sequences that trigger latent computation.  \Cref {fig:hybridexample} in Appendix~\ref{appendix:exmaplehybridCOT} presents an example  hybrid trajectory with $k_{\text{latent}}=1$.

We use $\tilde{s}_i$ to denote the symbolic-latent reasoning  step. Then the hybrid trajectory is
\begin{align}
\tilde{c} = (\tilde{s}_1,\tilde{s}_2 ,\cdots,\tilde{s}_N).
\end{align}
For ease of presentation, we also represent the hybrid trajectory in the form of concrete tokens, associated with binary labels whose entry indicates whether the  corresponding token is explicit:
\begin{equation}
\begin{aligned}
\label{eq:cgdef}
\tilde{c} &= (\tilde{c}_1, \tilde{c}_2, \cdots, \tilde{c}_{T}),  \\
\tilde{g} &= (\tilde{g}_1, \tilde{g}_2, \cdots, \tilde{g}_{T}),
\end{aligned}
\end{equation}
where $T$ is the total number of latent placeholders and explicit tokens, and $\tilde{g}_t=1$ indicates that $\tilde{c}_t$ is the explicit token.


\subsection{Model Architecture and Loss Function}
\label{sec:modelandloss}
We now introduce the model  architecture and  learning objective  based on the  hybrid trajectories.











\paragraph{Architecture.}  \method{} modifies the base  language model by incorporating a lightweight gating head, to decide whether the subsequent token should be verbalized. Specifically, given the hidden state $h_t$ at the last position, we compute a gate probability with  a two-layer MLP $p_{\rm{gate}}(t) = \sigma\!\left(W_2\,\mathrm{ReLU}(W_1 h_t)\right)$,
where $\sigma(\cdot)$ is the Sigmoid function. During inference, we obtain  a binary routing decision   $z_t = \mathbf{1}[p_{\rm{gate}}(t) \ge \tau]$ with $\tau$ a predefined threshold.  $z_t=1$ indicates that the next step is verbalized as  an explicit token, while $z_t=0$ indicates latent computation. Denote by $\hat{c}_{t+1}$  the generated  token, which is explicit when $z_t=1$.

The key mechanism of \method{} is that explicit and latent tokens correspond to different next-input representations.  When $z_t=1$, 
\method{} follows  standard autoregressive decoding and feeds the embedding of the generated token $\hat{c}_{t+1}$.  When $z_t=0$, it directly feeds back the current hidden state $h_t$ as the next-input embedding. Formally, the next-input embedding is given by
\begin{align}
\label{eqn:stateupate}
e_{t+1}=
\begin{cases}
e(\hat{c}_{t+1}), & z_t = 1,\\[4pt]
h_t, & z_t = 0.
\end{cases}
\end{align}

This design can automatically  decide when to verbalize intermediate  states to steer the reasoning process. It  distinguishes \method{} from prior latent reasoning methods such as  Coconut, CODI, and CoLaR, where
the number of latent thoughts is predetermined before  generating a response.

\paragraph{Supervised Loss Function.} We optimize a joint objective consisting of a language modeling loss and a gating loss. To facilitate the prediction of future reasoning,  the anchor  tokens and answer tokens are supervised, while latent placeholders are masked. Specifically, for a  sequence of hybrid tokens $\tilde{c}=\{\tilde{c}_t\}_{t=1}^{T}$  and binary labels $\tilde{g}=\{\tilde{g}_t\}_{t=1}^{T}$ defined in Eq.~(\ref{eq:cgdef}), the language modeling loss is
\begin{align}
\mathcal{L}_{\mathrm{LM}}
=
-\sum_{t} \tilde{g_t} \log p_{\theta} (\tilde{c}_t \mid \tilde{c}_{<t}),
\end{align}
where $\tilde{g_t}=1$ means that  $\tilde{c}_t$ is the explicit anchor token while  $\tilde{g_t}=0$ indicates a  latent placeholder.

In addition, we supervise the routing behavior at each hybrid token, involving the gating probability $p_{\mathrm{gate}}(t)$ and the binary label $\tilde{g}_t$:
\begin{align}
\mathcal{L}_{\mathrm{gate}}
=
\sum_{t}
\mathrm{CrossEntropy}\left(\tilde{g}_t, p_{\mathrm{gate}}(t) \right).
\end{align}

The final objective is then given by
\begin{align}
\label{eqn:totalloss2}
\mathcal{L}
=
\mathcal{L}_{\mathrm{LM}}
+
\lambda\mathcal{L}_{\mathrm{gate}},
\end{align}
where $\lambda>0$ is a balancing hyper-parameter.  This learning objective will be used in the following supervised training stage.


\subsection{Training and Inference}
We train \method{} in two phases: curriculum-based supervised training followed by RL training. During inference, \method{} performs gated autoregressive decoding in a mixed symbolic-latent space.

\paragraph{Curriculum-based Training.}

Inspired by the progressive curriculum strategy of Coconut \citep{hao2024coconut},  we adopt a  stage-wise curriculum that gradually transitions from  explicit CoT supervision to hybrid reasoning. Specifically, we construct a sequence of training trajectories with increasing degrees of hybridization. At stage~$k$, the first $k$ reasoning steps are converted into hybrid forms while remaining steps remain explicit:
\begin{align}
\tilde{c}^{(k)} = (\tilde{s}_1,\cdots,\tilde{s}_k,\; s_{k+1},\cdots,s_N).
\end{align}
Thus, stage~$0$ corresponds to the original explicit CoT $c$, while the final stage corresponds to the fully hybridized rationale $\tilde{c}$. 
At each stage, the model is trained on 
$\tilde{c}^{(k)}$
 using the supervised objective in Eq.~(\ref{eqn:totalloss2}). The hybrid reasoning tokens are additionally enclosed by two special boundary markers, \texttt{<start\_hybrid>} and \texttt{<end\_hybrid>}, allowing the model to distinguish hybrid reasoning from the question prompt and the final answer.

\paragraph{RL Training.}

We can further improve \method{}'s performance using Group Relative Policy Optimization (GRPO),  to explore  trajectories that are correct and more concise.
For each input question, we sample a group of $G$ hybrid reasoning trajectories and assign each trajectory  a reward $R_i$ combining response correctness and conciseness: 
\begin{align}
\label{eq:rlreward}
R_i
=
\mathbf{1}[\hat a_i=a]
\left[
1+\beta\left(1-\frac{T_i}{T_{\max}}\right)\right],
\end{align}
where \(\hat a_i\) and \(T_i\) are the generated answer and length of the $i$-th trajectory, respectively,  $a$ is the true answer, $\beta>0$ is a hyper-parameter, and $T_{\max}$ is a predefined maximum length.

We normalize the rewards $\{R_i\}_{i=1}^G$ to obtain group-relative advantages. At each step, we compute the importance ratio  over the sampled hybrid action. Unlike standard token-level policy ratios,  here the ratio includes both the gating and token-generation probabilities for anchor actions, but only the gating probability for a latent-computation action. GRPO then updates the policy model with  a clipped group-relative objective. Due to space limit, we provide the detailed formulation in \Cref{sec:rldetails}.


\paragraph{Inference Process.}
At inference, \method{} performs autoregressive generation in the hybrid space. The gating dynamically selects between generating an  explicit anchor token or continuing latent computation. In the explicit mode, the model generates the next token via standard decoding; in the latent mode, it feeds the current hidden state back as the next input and records a \texttt{<latent>} marker.  Once \texttt{<end\_hybrid>} is produced, the model exits  hybrid reasoning  and generates the final answer. In this way, \method{} can dynamically allocate its reasoning budget within  a unified autoregressive process.







\section{Experiments}
We evaluate the effectiveness of \method{} through experiments on mathematical reasoning tasks.

\begin{table*}[t]
    \centering
    \caption{
        Main results on in-domain and out-of-domain benchmarks.
        Performance is measured by accuracy (Acc., \%) and  average reasoning length (\# Tokens). 
    }
    \label{tab:main-results}
\vspace{-0.5em}
    \small
    \renewcommand{\arraystretch}{1.0}

    \resizebox{0.9\textwidth}{!}{%
    \begin{tabular}{@{}l*{10}{c}@{}}
        \toprule

        \multirow{3}{*}{Method}
        & \multicolumn{2}{c}{In-domain}
        & \multicolumn{6}{c}{Out-of-domain}
        & \multicolumn{2}{c}{Overall} \\

        \cmidrule(lr){2-3}
        \cmidrule(lr){4-9}
        \cmidrule(lr){10-11}

        & \multicolumn{2}{c}{GSM8K}
        & \multicolumn{2}{c}{GSM8K-Hard}
        & \multicolumn{2}{c}{Multi-Arith}
        & \multicolumn{2}{c}{SVAMP}
        & \multicolumn{2}{c}{Average} \\

        \cmidrule(lr){2-3}
        \cmidrule(lr){4-5}
        \cmidrule(lr){6-7}
        \cmidrule(lr){8-9}
        \cmidrule(lr){10-11}

        & Acc.(\%) & \#Tokens
        & Acc.(\%) & \#Tokens
        & Acc.(\%) & \#Tokens
        & Acc.(\%) & \#Tokens
        & Acc.(\%) & \#Tokens \\

        \midrule

        \multicolumn{11}{c}{GPT-2} \\
        \midrule

        CoT-SFT
        & 42.7 & 27.6 & 9.0 & 39.1 & 85.0 & 19.6 & 41.6 & 17.6 & 45.2 & 25.9 \\
        \midrule

        Coconut
        & 34.1 & 8.0 & 6.9 & 8.0 & 80.7 & 8.0 & 32.0 & 8.0 & 38.4 & 8.0 \\

        CODI
        & 24.6 & 8.0 & 5.0 & 8.0 & 45.0 & 8.0 & 26.9 & 8.0 & 25.4 & 8.0 \\

        CoLaR
        & 26.8 & 12.9 & 6.3 & 16.5 & 79.4 & 8.4 & 31.3 & 6.7 & 35.9 & 11.2 \\

        SIM-CoT
        & 33.1 & 12.0 & 7.5 & 12.0 & 81.6 & 12.0 & 32.6 & 12.0 & 38.7 & 12.0 \\

        \midrule

        {\method{}} (Ours)
        & 37.2 & 11.8 & 8.6 & 14.3 & 85.0 & 6.2 & 36.3 & 6.1 & 42.1 & 9.6 \\

        {\method-RL} (Ours)  &
{38.8} & 11.4 
& 8.4 & 13.5 
& {85.6}  & 6.8 
& {36.7} & 6.1 
&  42.4 & 9.4 \\
        \midrule

        \multicolumn{11}{c}{Llama-3.2-1B} \\
        \midrule

        CoT-SFT
        & 49.4 & 28.3 & 11.6 & 34.3 & 97.7 & 18.4 & 52.6 & 16.2 & 52.8 & 24.3 \\
        \midrule

        Coconut
        & 33.2 & 8.0 & 7.0 & 8.0 & 63.3 & 8.0 & 43.7 & 8.0 & 36.8 & 8.0 \\

        CODI
        & 37.1 & 8.0 & 9.3 & 8.0 & 90.6 & 8.0 & 47.2 & 8.0 & 46.1 & 8.0 \\

        CoLaR
        & 40.1 & 12.9 & 9.0 & 14.0 & 91.3 & 7.4 & 46.5 & 6.2 & 46.7 & 10.1 \\

        SIM-CoT
        & 39.0 & 12.0 & 9.1 & 12.0 & 86.7 & 12.0 & 42.5 & 12.0 & 44.3 & 12.0 \\

        \midrule

        {\method (Ours)}
        & 44.2 & 11.5 & 9.7 & 12.2 & 91.6 & 7.0 & 48.3 & 5.8 & 48.5 & 9.1 \\

        {\method-RL}  (Ours)
& {46.0}  & 11.4 
& 9.5 & 12.2 
& {92.2}  & 7.0
& {48.7}  & 5.9 
& {49.1} & 9.1 \\
        \bottomrule
    \end{tabular}%
    }
    \vspace{0.5em}
\end{table*}

\subsection{Experimental Setup}

\paragraph{Training Dataset.}
Unless otherwise stated, we train models on \textbf{GSM8K-Aug} \citep{deng2024gsm8k-aug}, a CoT augmented version of GSM8K \citep{cobbe2021gsm8k} that provides explicit intermediate reasoning trajectories with final answers.
Both  supervised training  and  subsequent RL training are conducted on this dataset, without using  additional data.

\paragraph{Evaluation Benchmarks.}
We report the experimental results on the \textbf{GSM8K-Aug} \citep{deng2024gsm8k-aug} test split, as our in-domain benchmark.
To evaluate the generalized performance beyond training distribution, we also test on three out-of-domain benchmarks:
(1) \textbf{SVAMP} \citep{patel2021SVAMP}, which evaluates robustness on arithmetic word problems with simple variations;
(2) \textbf{GSM8K-Hard} \citep{gao2023gsm8k-hard}, a harder GSM8K-style benchmark with increasing numerical magnitude and compositional difficulty;
and (3) \textbf{MultiArith} \citep{roy2015MultiArith}, a subset of MAWPS \citep{koncel2016mawps} consisting of multi-step arithmetic word problems.
See  \Cref{sec:datasets} for more dataset details.

\paragraph{Baselines.}
We consider the following baselines:
(1) \textbf{CoT-SFT}: finetunes the model on full CoT data and performs explicit token-level reasoning at inference;
(2) \textbf{Coconut} \citep{hao2024coconut}: gradually replaces explicit reasoning with latent reasoning through curriculum learning and performs reasoning via hidden-state feedback;
(3) \textbf{CODI} \citep{shen2025codi}:  compresses explicit CoT into the continuous latent space through self-distillation;
(4) \textbf{CoLaR} \citep{tan2025colar}: compresses multiple reasoning tokens into latent variables and further improves latent reasoning with dynamic compression;
(5) \textbf{SIM-CoT} \citep{wei2025sim}: adds step-level supervision via an auxiliary decoder.
\paragraph{Implementation.} 
For \method{}, we use the same hybrid trajectory construction throughout the paper: in each reasoning step we keep the top-$\gamma$ fraction of tokens as explicit anchors and compress the remaining tokens into latent placeholders, with $\gamma = 0.2$ and $k_\text{latent} = 1$.
Due to space limit, more implementation details and hyper-parameter settings are described in Appendix~\ref{sec:impl}.

\subsection{Main Results}

Table~\ref{tab:main-results} reports the main results on GPT-2 small and Llama-3.2-1B. 
 \method{} consistently outperforms prior fully latent  baselines in terms of accuracy while maintaining comparable or improved efficiency, demonstrating the effectiveness of verbalizing high-importance tokens as reasoning anchors. 
The RL-trained variant, \method-RL, further improves  performance and achieves the highest average accuracy among efficient reasoning methods. 

\begin{table*}[t]
    \centering
    \caption{
        Results on Qwen3-1.7B.
        Acc.(\%) denotes final-answer accuracy, and \#Tokens denotes
        average reasoning length, including both explicit tokens
        and latent steps.
    }
    \label{tab:acc-tokens}
\vspace{-0.5em}
    \small
    \renewcommand{\arraystretch}{1.0}

    \resizebox{0.9\textwidth}{!}{%
    \begin{tabular}{@{}l*{10}{c}@{}}
        \toprule

        \multirow{2}{*}{Method}
        & \multicolumn{2}{c}{GSM8K-NL}
        & \multicolumn{2}{c}{MATH}
        & \multicolumn{2}{c}{MATH500}
        & \multicolumn{2}{c}{ARC-C}
        & \multicolumn{2}{c}{Average} \\

        \cmidrule(lr){2-3}
        \cmidrule(lr){4-5}
        \cmidrule(lr){6-7}
        \cmidrule(lr){8-9}
        \cmidrule(lr){10-11}

        & Acc.(\%) & \#Tokens
        & Acc.(\%) & \#Tokens
        & Acc.(\%) & \#Tokens
        & Acc.(\%) & \#Tokens
        & Acc.(\%) & \#Tokens\\

        \midrule

        \multicolumn{11}{c}{Qwen3-1.7B} \\
        \midrule

        CoT-SFT
        & 69.5 & 81.8 & 35.6 & 321.1 & 35.0 
        & 302.7 & 57.6 & 541.7 & 49.4 & 311.8  \\






        \midrule

        HRPO
        & 64.5 & 110.2 & 27.6 & 763.2 & 26.0 
        & 765.1 & 53.7 & 464.2 & 43.0 & 525.7 \\

        Token Assorted
        & 60.1 & 73.9 & 26.9 & 183.4 & 27.6 
        & 253.4 & 66.0 & 88.3 & 45.2 & 149.8 \\

        \midrule

        {\method{} (Ours)}
        & 53.5 & 32.5 & 23.4 & 136.6 & 26.2 
        & 120.1 & 63.2 & 67.8 & 41.6 & 89.3\\

        {\method{}-RL (Ours)}
        & 62.5 & 37.7 & 34.4 & 140.3 & 34.8 
        & 136.5 & 71.6 & 70.3 & 50.8 & 96.2\\

        \midrule

    \end{tabular}%
    }
\end{table*}


Compared with CoT-SFT, \method-RL attains comparable average accuracy  with substantially shorter reasoning traces. It reduces the average reasoning length from 25.9 to 9.4 tokens on GPT-2 and from 24.3 to 9.1 tokens on Llama-3.2-1B. Notably, on MultiArith with GPT-2, \method-RL achieves an accuracy of 85.6\% with only 6.8 average tokens. 

\subsection{ Comparison with Hybrid Methods}
We further evaluate  \method{} against hybrid reasoning methods on more diverse and challenging benchmarks. Specifically, we train \method{} with Qwen3-1.7B on the training splits
of GSM8K-Aug-NL \citep{deng2024gsm8k-aug, shen2025codi} and MATH \citep{hendrycks2021MATH}, and  evaluate it on their test splits,
MATH-500 \citep{lightman2024MATH-500}, and ARC-C \citep{clark2018ARC}.
These experiments extend our evaluation beyond equation-centric GSM8K-Aug
setting to  natural-language mathematical reasoning, more challenging mathematical problems,  and science reasoning.
We compare \method{} with CoT-SFT, HRPO \citep{yue2026hybrid},
and Token Assorted \citep{su2025mix}, reporting  answer
accuracy and  total reasoning length, which includes 
explicit tokens and latent steps.

For \method{}-RL,
we use a  binary correctness reward
$R_i = \mathbf{1}[\hat a_i = a]$. We omit the length penalty to avoid explicitly favoring shorter reasoning trajectories in this comparison, since these two hybrid methods generally have much longer reasoning trajectories than fully latent methods.

As shown in Table~\ref{tab:acc-tokens}, 
\method{} produces the shortest trajectories, albeit with lower average accuracy than the baselines.
RL training substantially improves accuracy across all four benchmarks, increasing average accuracy from 41.6\% to 50.8\%, with only  7.7\% increase in average reasoning length (from 89.3 to 96.2).
Compared with Token Assorted, \method{}-RL achieves higher accuracy with shorter trajectories on all four benchmarks. It also outperforms HRPO in terms of accuracy on three of the four benchmarks while maintaining shorter trajectories on all the benchmarks.

\begin{figure}[t]
    \centering
    \includegraphics[width=1\columnwidth]{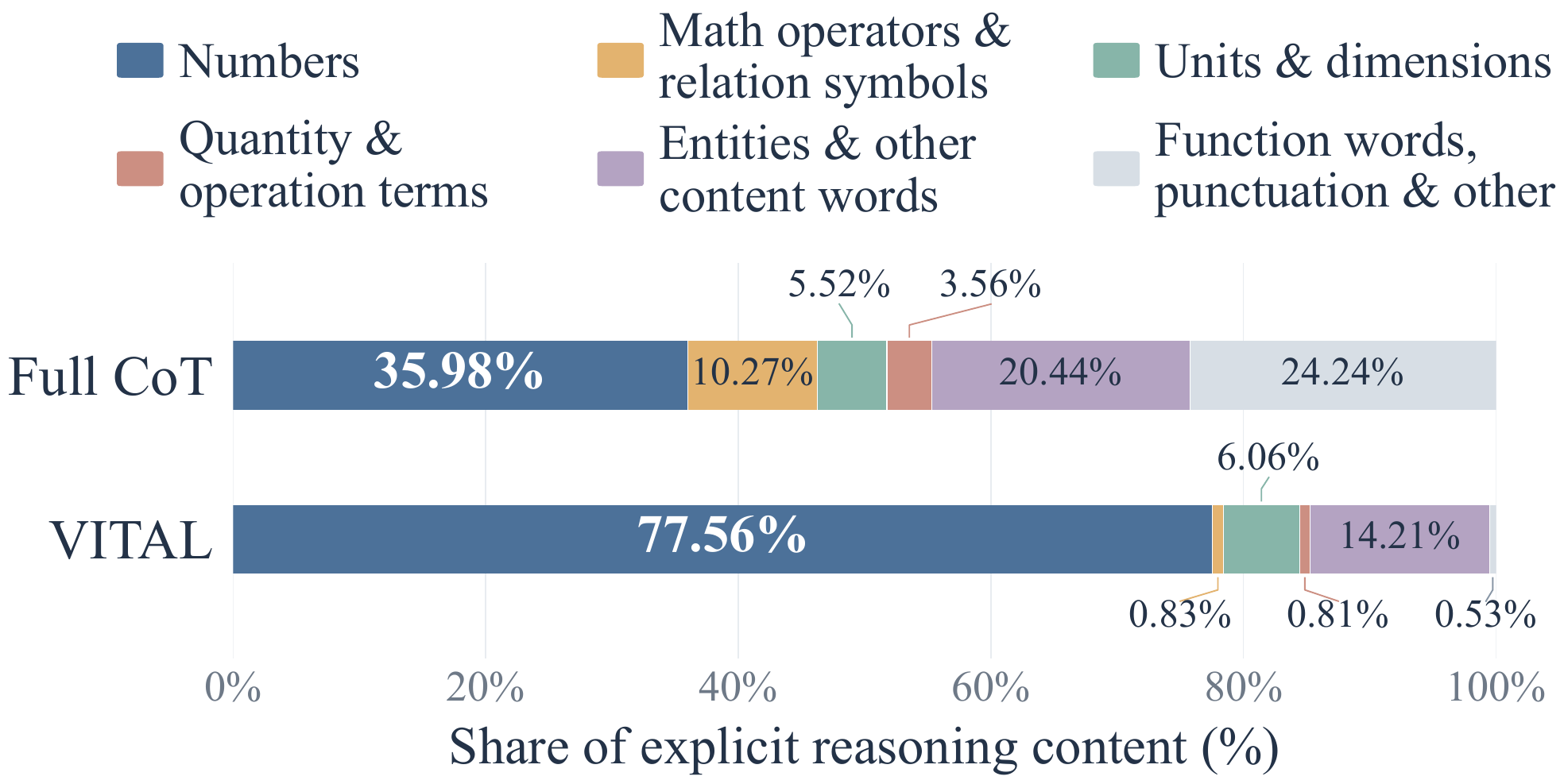}
        \caption{Token-category distributions of full CoT and \method{} on GSM8K-NL (Qwen3-1.7B). }
    \label{fig:token-composition}
\end{figure}

\subsection{Analysis of Generated Tokens}
\label{sec:anchor-analysis}


To better understand what \method{} selectively verbalizes,  
 we compare the token-category distributions of its explicit outputs on Qwen3-1.7B with those of full CoT.
As shown in Figure~\ref{fig:token-composition}, numerical tokens account for a substantially larger proportion of \method{}'s explicit outputs, while function words, punctuation, and other general textual content are less prevalent.
This suggests that \method{} does not merely generate shortened natural language rationales, but instead prioritizes compact numerical and symbolic information in its explicit outputs. 


\subsection{Robustness to Error Accumulation}

\begin{figure}[t]
    \centering
    \includegraphics[width=0.9\columnwidth]{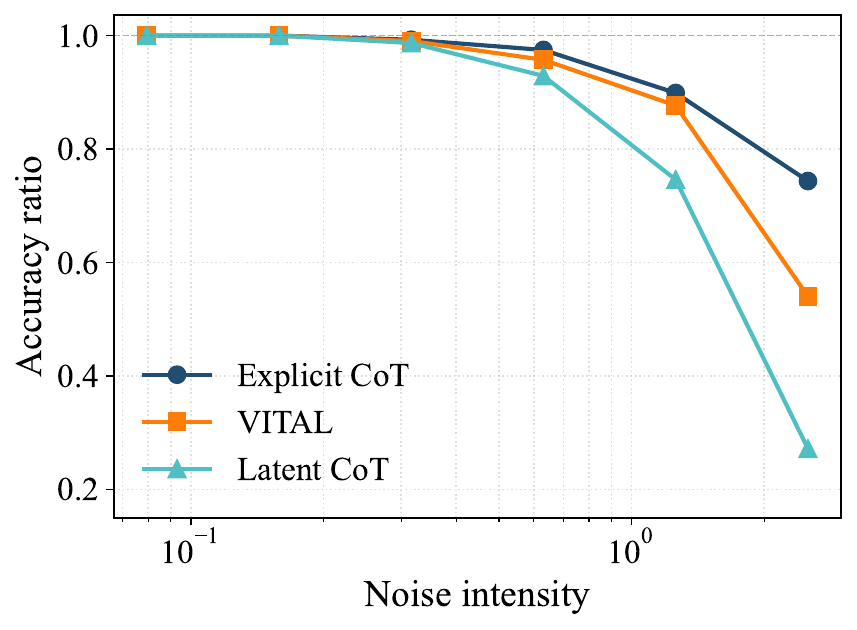}
    \vspace{-0.5em}
        \caption{Performance degradation under noise injection. We inject Gaussian noise $\mathcal{N}(0, \sigma^2 I_d)$  to the hidden states during reasoning and report the accuracy result of GSM8K on Llama-3.2-1B. }
    \label{noise1}
\end{figure}

 As discussed in \citet{zou2026capacityandlimits},  small errors can accumulate when the continuous states are repeatedly propagated in fully latent reasoning methods.  In contrast, explicit CoT  is more robust  as it discretizes intermediate states into concrete tokens. To evaluate the robustness of \method{} against error accumulation, we use a similar setting from \citet{zou2026capacityandlimits} and  inject Gaussian noise into the hidden states during the reasoning process. \Cref{noise1} reports the performance degradation under different noise intensity on Llama-3.2-1B. In comparison with the fully latent CoT method Coconut,  emitting  explicit tokens as reasoning anchors clearly improves the  robustness to noises, validating the effectiveness of \method{}.

\subsection{Ablation Study}

\paragraph{Anchor Selection.} To further investigate  the contribution of important tokens  to the performance gains of \method{},  we compare three token-selection strategies using
Qwen3-1.7B: highest-importance selection (used by \method{});
random selection, and lowest-importance selection.

Table~\ref{tab:anchorablation} reports the accuracy results on   GSM8K-Aug, GSM8K-Hard, MultiArith, and SVAMP.
Highest-importance selection outperforms both alternatives
on all four benchmarks.
Random selection obtains smaller average gains over Coconut, whereas
the performance of lowest-importance selection performs is close to Coconut.
This result highlights the effectiveness of importance-guided anchor selection, suggesting that the choice of verbalized tokens matters more than merely introducing explicit tokens.

\begin{table}[t]
\centering
\caption{Accuracy (\%) on four benchmarks. \method{} uses highest-importance tokens as reasoning anchors.}
\vspace{-0.5em}
\label{tab:anchorablation}
\setlength{\tabcolsep}{3pt}
\renewcommand{\arraystretch}{1.0}
\small
\resizebox{\columnwidth}{!}{%
\begin{tabular}{lccccc}
\toprule
Method & GSM8K-Aug & GSM8K-Hard & MultiArith & SVAMP & Avg. \\
\midrule
SFT                  & 69.1  & 36.5 & 98.9 & 77.5 & 71.3 \\
\midrule 
Coconut              & 60.2 & 25.3 & 97.2 & 65.4 & 62.0 \\

\midrule
Lowest    & 59.2 & 25.1 & 97.8 & 66.1 & 62.1 \\
Random               & 62.8 & 28.2 & 97.8 & 67.8 & 64.1 \\

\method{}      & {64.1} & {30.1} & {98.3} & {69.2} & {65.4} \\
\bottomrule
\end{tabular}
}\vspace{1em}
\end{table}

\begin{figure}[t]
    \centering
    \includegraphics[width=0.94\columnwidth,height=0.7\columnwidth]{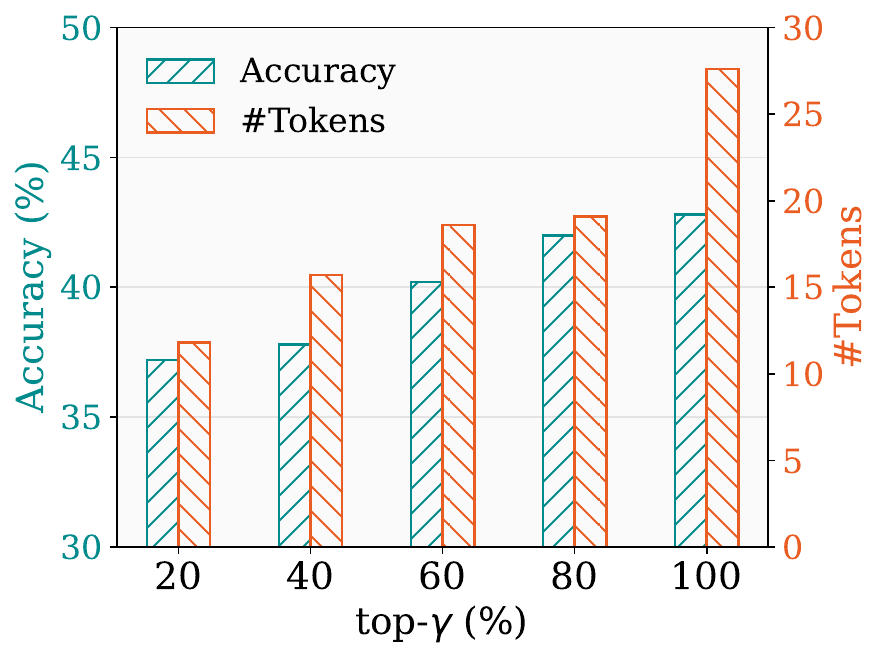}\vspace{-0.5em}
    \caption{Ablation on the top-$\gamma$ ratio. We vary the proportion of explicit  anchors and evaluate its effect.}
    \label{fig:top-gamma}
\end{figure}

\paragraph{Top-$\gamma$ Ratio.} We study how the top-$\gamma$ ratio in hybrid trajectory construction affects reasoning performance.  Specifically, we vary $\gamma$ in \{0.2, 0.4, 0.6, 0.8, 1.0\}, where larger values retain more tokens as reasoning anchors and smaller values enable greater latent compression.  
As shown in Figure~\ref{fig:top-gamma}, increasing $\gamma$ generally improves accuracy, as more  information is preserved in explicit anchors. However, this also increases the reasoning length.
Overall, $\gamma=0.2$  appears to achieve the best trade-off, retaining sufficient anchors for stable reasoning while keeping most computation latent.

\section{Concluding Remarks}
We propose \method{}, a hybrid reasoning framework that learns when to think aloud and when to remain silent. \method{} selectively verbalizes  informative tokens  as  anchors for latent reasoning. It introduces a lightweight gating head to the base language model. For training, we construct hybrid  trajectories from explicit rationales, followed by a two-stage training process consisting of curriculum-based supervised training and RL training. Experiments show that \method{} consistently outperforms existing fully latent reasoning methods at comparable reasoning lengths. Compared with hybrid baselines, \method{} achieves higher reasoning accuracy with substantially  shorter reasoning trajectories. These results validate the effectiveness of importance-guided selective verbalization in improving the accuracy-length trade-off of latent reasoning.

\newpage
\section*{Limitations}
In this paper, the largest model on which we conduct supervised training is Llama-3.2-3B-Instruct (see Appendix~\ref{sec:appendix-llama3b}). Due to
 limited GPU resources, we are not able to complete RL training at this scale, and our RL results are limited to smaller models. Another limitation is on the choice of reasoning anchors. The present work considers only the ``important'' tokens in the original explicit CoT trajectories, and the proposed method is effective in improving  the accuracy-length trade-off for latent reasoning. Yet, we do not investigate  other forms of reasoning anchors. Future work can explore suitable hidden embeddings or other supervision signals as the reasoning anchors.
 \section*{Ethics Impact}
The proposed method \method{} is designed to improve both reasoning performance and efficiency.  Part of the reasoning process is performed in latent space, and the generated rationales may expose only a few explicit anchor tokens rather than the full internal reasoning trajectory. This may make model behavior less transparent than fully explicit CoT in some settings. Therefore, \method{} should be used with appropriate task-specific evaluation and safety checks before deployment in high-stakes applications. We will also release our
 code to facilitate future research on efficient and transparent LLM reasoning.

\bibliography{custom}

\clearpage
\appendix

\section{RL Training Details}
\label{sec:rldetails}
Following the reward defined in  Eq.~(\ref{eq:rlreward}), we estimate the advantage through group-relative reward normalization. The advantage of the $i$-th trajectory is  given by
\[
A_i = \frac{R_i-\mathrm{mean}(\{R_i\}_{i=1}^G)}
{\mathrm{std}(\{R_i\}_{i=1}^G)}.
\]
At each step $t$, we compute the importance ratio  over the sampled hybrid action. Let $z_{i,t}$ denote the binary decision of the gating function in the $i$-th trajectory and $\hat{c}_{i,t}$ the explicit token with $z_{i,t}=1$. Then we  define the hybrid action as:
\[
y_{i,t} =
\begin{cases}
(z_{i,t}, \hat{c}_{i,t}), & z_{i,t}=1, \\
z_{i,t}, & z_{i,t}=0.
\end{cases}
\]
The policy probability of a hybrid action depends on whether the action corresponds to explicit anchor generation or latent computation, which is
\begin{align}
&\pi_{\theta}(y_{i,t} \mid x, y_{i,<t}) \nonumber\\ &=
\begin{cases}
p_{\mathrm{gate},\theta}(t)
p^{\mathrm{LM}}_{\theta}(\hat{c}_{i,t} \mid x, y_{i,<t}),
& z_{i,t}=1, \\
1 - p_{\mathrm{gate},\theta}(t),
& z_{i,t}=0, \nonumber
\end{cases}
\end{align}
and the importance ratio is defined as
\[
\rho_{i,t}
=
\frac{
\pi_\theta(y_{i,t}\mid x,y_{i,<t})
}{
\pi_{\theta_{\mathrm{old}}}(y_{i,t}\mid x,y_{i,<t})
},
\]
where $\pi_\theta$ and $\pi_{\theta_{\mathrm{old}}}$ are the current and old policy models, $x$ is the input question, $y_{i,t}$ is the hybrid action corresponding to either anchor generation or latent computation, and $y_{i,<t}$ is the $i$-th trajectory before the $t$-th step. GRPO then minimizes the clipped policy loss:
\begin{align}
\mathcal{L}_{\mathrm{GRPO}}
=
-\frac{1}{G}
&\sum_{i=1}^{G}
\sum_{t=1}^{T_i}
\min\Big(
\rho_{i,t}\frac{A_i}{T_i}, \nonumber\\
&
\hspace{20pt} \mathrm{clip}(\rho_{i,t},1-\epsilon,1+\epsilon)\frac{A_i}{T_i}
\Big), \nonumber
\end{align}
where $\epsilon$ is a clipping hyperparameter and $T_i$ is the length of the $i$-th reasoning trajectory. Here we also remove the KL-regularization term from original GRPO for efficient training, following \citet{yu2025dapo,tan2025colar}.


\section{Datasets}
\label{sec:datasets}
We provide examples and statistics of the training datasets and evaluation benchmarks.
\subsection{Examples}
\begin{datasetbox}{GSM8k-Aug}
Question = "Out of 600 employees in a company, 30\% got
promoted while 10\% received bonus. How many employees did
not get either a promotion or a bonus?"\\
CoT = "<<600*30/100=180>><<600*10/100=60>>\\
<<180+60=240>><<600-240=360>>"\\
Answer = "360"
\end{datasetbox}

\begin{datasetbox}{GSM-Hard}
Question = "Marie ordered one chicken meal that costs \$12, 5 packs of milk that costs \$3 each, 4 apples that cost \$1.3299286 each, and some boxes of pizza. Marie paid a total of \$3299286. How many boxes of pizza did Marie order if each box costs \$8.3299286?"\\
Answer = "388147.4117647059"
\end{datasetbox}

\begin{datasetbox}{MultiArith}
Question = "A waiter at 'The Greasy Spoon' restaurant had 26 customers to wait on. During the lunch rush he added another 27 customers. If 27 of the customers didn't leave him a tip, how many customers did leave a tip?"\\
Answer = "26"
\end{datasetbox}

\begin{datasetbox}{SVAMP}
Question = "Paul got a box of some crayons for his birthday. During the school year he gave 52 crayons to his friends while he lost 535 crayons. If he only had 492 crayons left, how many crayons had been lost or given away?"\\
Answer = "587.0"
\end{datasetbox}

\subsection{Dataset Statistics}
The statistics of the training data are shown in
Table~\ref{tab:train-stats}, and the statistics of the
evaluation benchmarks are shown in Table~\ref{tab:eval-stats}.

\begin{table}[h]
\centering
\small   
\caption{Training data statistics.\vspace{-0.5em}}
\begin{tabular}{l | c c}
\toprule
\textbf{Training Dataset} & Num. Data & Avg. CoT Tokens \\
\midrule
GSM8K-Aug & 385,620 & 20.3 \\
\bottomrule
\end{tabular}
\label{tab:train-stats}
\end{table}

\begin{table}[h]
\centering
\small   
\caption{Evaluation benchmark statistics. \vspace{-0.5em}}
\begin{tabular}{l | c c}
\toprule
\textbf{Benchmark} & Type & Data Size \\
\midrule
GSM8K-Aug   & In-domain  & 1,319 \\
GSM8K-Hard  & Out-of-domain & 1,319 \\
SVAMP       & Out-of-domain & 1,000 \\
MultiArith  & Out-of-domain & 500   \\
\bottomrule
\end{tabular}
\label{tab:eval-stats}
\end{table}

\section{Implementation Details}
\label{sec:impl}

\subsection{General Setup}
We evaluate \method{} on three backbones: GPT-2, Llama-3.2-1B, and Llama-3.2-3B.
{All models are trained solely on GSM8K-Aug.
To ensure a fair comparison, all compared methods adopt the same training dataset and evaluation
protocol.}
Due to limited computing resources,    each dataset-level result is obtained from a single run  unless otherwise stated. GPT-2 \citep{radford2019gpt2} and Llama-3.2-1B are trained in fp32 precision,
while Llama-3.2-3B \citep{meta2024llama32} is trained in bf16 precision.

We measure reasoning efficiency by the average reasoning chain length (\# Tokens).
We count only the tokens within the reasoning chain, i.e., the explicit anchor tokens
together with the \texttt{<latent>} placeholders, and exclude the question and the final answer tokens.

\subsection{Setup of \method{}}
\paragraph{Hybrid Trajectory Construction.}
We employ LLMLingua-2~\citep{pan-etal-2024-llmlingua} as a frozen token-importance scoring module.
Within each reasoning step, we retain the top-$\gamma$ fraction of tokens as explicit anchors
with $\gamma=0.2$, and compress each segment of consecutive non-anchor tokens into
$k_{\text{latent}}=1$ \texttt{<latent>} placeholder.
The hybrid reasoning tokens are enclosed by the special markers
\texttt{<start\_hybrid>} and \texttt{<end\_hybrid>}.

\paragraph{Gating and Loss.}
At inference, a token is generated explicitly when the gate probability exceeds the
threshold $\tau=0.5$.
The training objective combines the language-modeling loss and the gating loss with a
balancing weight $\lambda=0.1$.

\paragraph{RL Training.}
After supervised training, we further optimize \method{} with GRPO.
We pick group size  $G=32$,  conciseness weight $\beta=0.5$ for the reward,  clipping
parameter $\epsilon=0.2$, and a learning rate of $5\times10^{-6}$.

\subsection{Setup of Baseline Methods}
We reproduce the baseline methods following their default configurations
reported in the papers.

\paragraph{Coconut}\citep{hao2024coconut}: 
We use a fixed learning rate of $1\times10^{-4}$ and represent each reasoning step with
$2$ continuous thoughts. A multi-stage curriculum progressively converts explicit
reasoning steps into latent thoughts. The model is trained for $3$ epochs in each  stage except the last stage.

\paragraph{CODI}\citep{shen2025codi}:
We use the AdamW optimizer with a cosine learning-rate schedule, linear warm-up over the first
$3\%$ training steps, an effective batch size of $128$. The model uses $6$ continuous thoughts, and the distillation objective
weights are set to be $1$ for both the student and distillation terms. The learning
rate is $3\times10^{-3}$ with $40$ epochs for GPT-2 and $8\times10^{-4}$ with $10$ epochs
for Llama-3.2-1B.

\paragraph{CoLaR}\citep{tan2025colar}:
We use the AdamW optimizer with a weight decay of $10^{-2}$ and a learning rate of
$1\times10^{-4}$ for supervised fine-tuning. The latent head is a three-layer MLP, the maximum compression
factor is  $5$, and the finetuning batch size is $256$. 

\paragraph{SIM-CoT}\citep{wei2025sim}:
SIM-CoT is a plug-and-play step-level supervision module that introduces an auxiliary
decoder during training, which is removed at inference. It is applied on top of Coconut
and CODI, inheriting their respective configurations above. We use Adam
($\beta_1=0.9$, $\beta_2=0.999$ and weight decay $0.1$) optimizer and set batch size to $128$ for GPT-2 and
Llama-3.2-1B.

\subsection{Summary of Hyperparameters}
Table~\ref{tab:voice-hparams} summarizes the key hyperparameters of \method{}, and
Table~\ref{tab:baseline-hparams} summarizes the configurations of  baselines.

\begin{table}[h]
\centering
\small
\caption{Hyperparameters of \method{}.\vspace{-0.5em}}
\begin{tabular}{l | l}
\toprule
\textbf{Hyperparameter} & \textbf{Value} \\
\midrule
Anchor ratio $\gamma$            & 0.2 \\
Latent placeholders $k_{\text{latent}}$ & 1 \\
Gating threshold $\tau$          & 0.5 \\
Loss weight $\lambda$            & 0.1 \\
Curriculum stages                & 3 \\
Epochs per stage                 & 3 \\
RL group size $G$                & 32 \\
RL reward coefficient $\alpha$   & 0.5 \\
RL clipping $\epsilon$           & 0.2 \\
RL learning rate                 & $5\times10^{-6}$ \\
\bottomrule
\end{tabular}
\label{tab:voice-hparams}
\end{table}

\begin{table}[h]
\centering
\small
\caption{Configurations of the baselines. All models are trained for the specified number of epochs or 50 hours, whichever is reached first. \vspace{-0.5em}}
\begin{tabular}{l | l c c}
\toprule
\textbf{Method} & \textbf{Model} & \textbf{LR} & \textbf{Epochs} \\
\midrule
\multirow{3}{*}{Coconut} & GPT-2 & $1\times10^{-4}$ & 20 \\
                         & Llama-3.2-1B  & $8\times10^{-4}$ & 20 \\
                         & Llama-3.2-3B  & $8\times10^{-4}$ & 20 \\
\midrule
\multirow{3}{*}{CODI} & GPT-2     & $3\times10^{-3}$ & 40 \\
                      & Llama-3.2-1B  & $8\times10^{-4}$ & 10 \\
                      & Llama-3.2-3B  & $8\times10^{-4}$ & 10 \\
\midrule
\multirow{3}{*}{CoLaR} & GPT-2     & $1\times10^{-4}$ & 50 \\
                      & Llama-3.2-1B  & $1\times10^{-4}$ & 50 \\
                      & Llama-3.2-3B  & $1\times10^{-4}$ & 50 \\
\bottomrule
\end{tabular}
\label{tab:baseline-hparams}
\end{table}

\paragraph{Computing Budget.}
All the experiments are conducted on a compute node with 2 NVIDIA H800 80GB GPUs. For supervised training, the GPT-2, Llama-3.2-1B, and Llama-3.2-3B experiments require approximately 6, 40, and 100 hours, respectively.

\section{An Example of Constructed Hybrid CoT.} 
\label{appendix:exmaplehybridCOT}

\Cref {fig:hybridexample} shows an example  hybrid trajectory with $k_{\text{latent}}=1$.

\begin{figure}[h]
\centering
\begin{datasetbox}{\normalsize Example of Hybrid Trajectory}
\small
\rmfamily

\textbf{Problem:}
A car travels with a speed of 80 miles per hour. If it has already travelled  6 hours, how long more does it have to travel if the total distance is 1000 miles?

\vspace{0.4em}

\textbf{Original CoT:}
\vspace{-0.3em}
\begin{flushleft}
\ttfamily
\textless\textless 80 * 6 = 480\textgreater\textgreater\\
\textless\textless 1000 - 480 = 520\textgreater\textgreater\\
\textless\textless 520 / 80 = 6.5\textgreater\textgreater
\end{flushleft}


\textbf{Hybrid CoT:}
\vspace{-0.3em}
\begin{flushleft}
\ttfamily
\textcolor{red!75!black}{\textless latent\textgreater}
* 6
\textcolor{red!75!black}{\textless latent\textgreater}\\
\textcolor{red!75!black}{\textless latent\textgreater}
- 480
\textcolor{red!75!black}{\textless latent\textgreater}\\
\textcolor{red!75!black}{\textless latent\textgreater} 80
\textcolor{red!75!black}{\textless latent\textgreater}
6.5
\end{flushleft}


\textbf{Answer:} 6.5

\end{datasetbox}
\vspace{-0.5em}
\caption{An example of the constructed hybrid CoT.}
\label{fig:hybridexample}
\end{figure}

\section{Additional Results on Larger Backbones}
\label{sec:appendix-llama3b}

To examine whether \method{} remains effective on a larger model, we consider  Llama-3.2-3B with the same GSM8K-Aug setting as in
\Cref{tab:main-results} and compare \method{} against CoT-SFT and the latent
baselines Coconut, CODI, and CoLaR. Due to limited computational
resources, we report results for the supervised model only and do not
perform RL training at this scale. The results are reported in
\Cref{tab:llama-main-results-full}.

On Llama-3.2-3B, \method{} continues to provide a favorable
accuracy-length trade-off. It achieves 54.1\% average accuracy with
8.3 reasoning steps, outperforming the strongest latent baseline,
CODI, by 2.1 percentage points at a similar reasoning length.
Compared with CoLaR, it improves average accuracy by 3.0 percentage
points while reducing the average reasoning length from 16.8 to 8.3.
Compared with CoT-SFT, \method{} is only 2.4 percentage points lower
in average accuracy while using 34.6\% of its reasoning length.

\begin{table*}[htbp]
\caption{Experimental results on Llama-3.2-3B. Accuracy and average reasoning chain length are reported across in-domain and out-of-domain benchmarks. \vspace{-0.5em}}
\centering
\renewcommand{\arraystretch}{1.05}
\resizebox{0.9\textwidth}{!}{%
\begin{tabular}{l cc cc cc cc cc}
\toprule

\multirow{3}{*}{\textbf{Method}} 
& \multicolumn{2}{c}{\textbf{In-domain}} 
& \multicolumn{6}{c}{\textbf{Out-of-domain}} 
& \multicolumn{2}{c}{\textbf{Overall}} \\
\cmidrule(lr){2-3} \cmidrule(lr){4-9} \cmidrule(lr){10-11}

& \multicolumn{2}{c}{GSM8K-Aug} 
& \multicolumn{2}{c}{GSM8K-Hard} 
& \multicolumn{2}{c}{MultiArith} 
& \multicolumn{2}{c}{SVAMP} 
& \multicolumn{2}{c}{Average} \\
\cmidrule(lr){2-3} \cmidrule(lr){4-5} \cmidrule(lr){6-7} 
\cmidrule(lr){8-9} \cmidrule(lr){10-11}

& Acc. (\%) & \# Tokens 
& Acc. (\%) & \# Tokens 
& Acc. (\%) & \# Tokens 
& Acc. (\%) & \# Tokens 
& Acc. (\%) & \# Tokens \\
\midrule
 CoT-SFT 
& 55.4 & 28.5 
& 12.7 & 35.7 
& 96.7 & 17.6 
& 61.0 & 14.5 
& 56.5 & 24.0 \\
\midrule

Coconut 
& 36.7 & {8.0} 
& 8.7 & {8.0} 
& 90.5 & 8.0 
& 37.8 & 8.0 
& 43.4 & {8.0} \\

CODI
& {47.2} & {8.0} 
& {11.2} & {8.0} 
& 93.2 & 8.0 
& {56.5} & 8.0 
& 52.0 & {8.0} \\

CoLaR 
& 45.4 & 13.5
& 10.1 & 40.4
& {96.3} & 7.3 
& 51.8 & 6.0 
& 51.1 & 16.8 \\
\midrule

 \method{} 
& {52.2} & 11.9 
& {11.6} & 10.2 
& {97.2} & {6.1}
& 55.2 & {5.1} 
& {54.1} & 8.3 \\

\bottomrule
\end{tabular}%
}

\label{tab:llama-main-results-full}
\end{table*}

\section{Artifact Licenses and Responsible Use}
\label{sec:artifact-responsible-use}

\paragraph{Artifacts.}
This work uses existing public datasets, pretrained models, and baseline methods.
The datasets include GSM8K-Aug, GSM8K-Hard, MultiArith, SVAMP, and CommonsenseQA.
The pretrained backbones include GPT-2 and Llama-3.2 models.
We also use LLMLingua-2 as a frozen token-importance scorer and compare our method with prior latent reasoning baselines including Coconut, CODI, CoLaR, and SIM-CoT.

\paragraph{Licenses and Terms of Use.}
We use these artifacts for research and benchmarking purposes.
We check the licenses and terms of use of the public artifacts used in our experiments.
GSM8K is released under the MIT License, GPT-2 is released under a modified MIT License, and CommonsenseQA is distributed under the MIT License according to its dataset card.
Llama-3.2 models are used under the Llama-3.2 Community License and its acceptable use policy.
For all artifacts, we follow their intended research use and do not redistribute any restricted model weights or data beyond what is permitted by the corresponding licenses.

\paragraph{Intended Use.}
The intended use of our experiments is to study efficient reasoning in language models.
The derived models and experimental results are intended for research evaluation only.
They should not be used directly in high-stakes domains such as medical, legal, financial, or safety-critical decision making without additional validation and safeguards.

\paragraph{Privacy and Offensive Content.}
We do not collect new user data and do not conduct human-subject data collection.
The datasets used in this work are existing public benchmarks.
They consist primarily of mathematical word problems and multiple-choice commonsense questions.
We do not intentionally use personally identifying information.
However, as with many public NLP datasets, we cannot completely rule out the presence of unintended sensitive or biased content.
We therefore restrict our use of these datasets to research evaluation and report aggregate benchmark results only.

\end{document}